\documentclass{article}
\usepackage{spconf, amsmath, graphicx}
\usepackage[colorlinks,
            linkcolor=blue,
            anchorcolor=blue,
            citecolor=blue]{hyperref}
\usepackage[table]{xcolor}
\newcommand{\gr}{\rowcolor[gray]{.95}}
\usepackage{multirow}
\title{SANDFORMER: CNN and Transformer under Gated Fusion for Sand Dust Image Restoration}
\name{Jun Shi$^{1\dag}$, Bingcai Wei$^{2\dag}$, Gang Zhou$^{1*}$, Liye Zhang$^{2}$\thanks{$^{*}$Corresponding author. $\dag$: Equal Contribution. Email: junshi2022@gmail.com. This work was supported in part by the National Natural Science Foundation of China under Grant 62166040, Grant 62261053, Grant 62137002, and in part by the Natural Science Foundation of XinJiang under Grant 2021D01C057.}}
\address{$^{1}$School of Information Science and Engineering, Xinjiang University\\$^{2}$College of computer science and technology, Shandong University of Technology}

\begin{document}
%\ninept
%
\maketitle
\begin{abstract}
Although Convolutional Neural Networks (CNN) have made good progress in image restoration, the intrinsic equivalence and locality of convolutions still constrain further improvements in image quality. Recent vision transformer and self-attention have achieved promising results on various computer vision tasks. However, directly utilizing Transformer for image restoration is a challenging task. In this paper, we introduce an effective hybrid architecture for sand image restoration tasks, which leverages local features from CNN and long-range dependencies captured by transformer to improve the results further. We propose an efficient hybrid structure for sand dust image restoration to solve the feature inconsistency issue between Transformer and CNN. The framework complements each representation by modulating features from the CNN-based and Transformer-based branches rather than simply adding or concatenating features. Experiments demonstrate that SandFormer achieves significant performance improvements in synthetic and real dust scenes compared to previous sand image restoration methods.
\end{abstract}

% The key insight of this study is to investigate how to combine CNN and Transformer for sand image restoration. 
\vspace{-0.5cm}
\section{Introduction}

Sandstorms are one of the most common dynamic weather phenomena and can significantly reduce the visibility and contrast of captured images. The existing sand dust degraded image restoration methods are mainly based on the atmospheric light scattering model to improve the classic haze removal algorithm directly. Due to some assumptions and prior knowledge constraints, the processing results of sand dust images will have color cast and blur. In recent years, the use of deep learning methods to enhance haze images has achieved great success. Inspired by this, some scholars began to apply deep learning methods to sand dust image restoration. 

% Experiments show that there are many problems in directly applying the deep learning-based image dehazing method to sand dust image restoration, such as color cast, artifacts, and serious loss of details. Therefore, it is crucial to construct a restoration network for sand dust images.

% \begin{figure}[!h]
%     \centering
%     \includegraphics[width=1\linewidth]{Images/fig1_1.pdf}
%     \caption{Visual comparison of our method with state-of-the-art sand dust image restoration methods such as MSBDN\cite{dong2020multi} and \cite{2021TransWeather} on real sand dust images. The results of \cite{dong2020multi} may be difficult to remove color shifts, and the methods in \cite{2021TransWeather} are difficult to preserve full details. In contrast, the results of our method are sharper and richer in detail.}
%     \label{fig1}
% \end{figure}

\cite{shizhenghao2022} proposed a convolutional neural network sand dust image enhancement method with color restoration. \cite{si2022comprehensive} proposed a sand dust image reconstruction benchmark for training convolutional neural networks and evaluating the algorithm's performance, using the existing Pix2Pix network to restore sand dust images. However, the CNN-based architecture only considers the local features of the image, making the overall color projection problem of the image difficult to solve. At the same time, there are certain limitations in capturing long-range dependencies and recovering weak texture details. Based on this, vision transformer (ViT) \cite{dosovitskiy2020image, touvron2021training} came into being. ViT demonstrated the advantage of global processing and achieved a significant performance boost over CNN. Transformer can provide long-distance feature dependencies via the cascaded self-attention. However, it lacks the capability of retaining local feature details, thus leading to ambiguous and coarse details for image reconstruction. Therefore, it is very important to effectively combine CNN and Transfoemr.

%The Transformer \cite{vaswani2017attention} has recently gained particular attention in the natural language processing (NLP) community thanks to its impressive performance. The key component contributing to the success is a self-attention mechanism, which beneficially captures long-range dependencies among input sequences \cite{yoo2022rich}.

% Combining convolutional and Transformer layers can achieve better model generalization and efficiency.

\begin{figure*}[!h]
    \centering
    \includegraphics[width=1\linewidth]{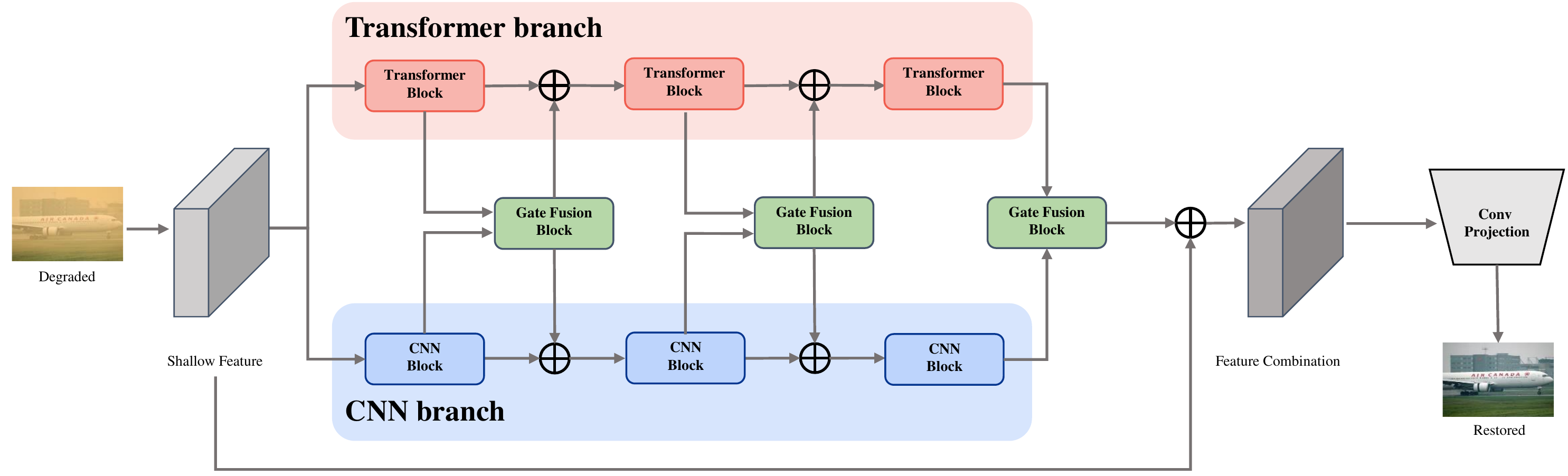}
    \caption{Overview structure of our method.}
    \label{framework_1}
\end{figure*}

Motivated by this, we propose a new design that brings together the power of Transformer and CNN into sand image restoration. The main idea is illustrated in Figure \ref{framework_1}. Specifically, we first introduce an effective hybrid architecture that takes advantage of CNN and recent ViT for sand image restoration. We propose two branches (i.e., CNN and transformer branches) and aggregate them several times during the image restoration procedure. Consequently, local features extracted from the CNN branch and long-range dependencies captured in the transformer branch are progressively fused to complement each other and extract rich features. Experiments and comparisons demonstrate the superiority of our method over state-of-the-art sand image restoration methods. 

In summary, our contributions are presented as follows:
\begin{itemize}
    \item  In comparison to pure CNN-based sand image restoration networks, our work is the first to introduce the power of Transformer into sand image restoration via novel designs.
    \item We propose a new fusion method to fuse CNN-based features with Transformer backbone-based features effectively.
    \item Extensive experiments on SandPascal VOC++ and real image datasets demonstrate the excellent performance of our method.
\end{itemize}

\vspace{-0.5cm}
\section{Method}

In this section, we detail how to leverage the respective strengths of CNN and Transformer to promote their property in image restoration tasks to restore sharper images. SandFormer consists of a shallow feature extraction module, a transformer branch, a CNN branch, and a high-quality projected image restoration module. The overall structure of SandFormer is shown in Figure \ref{framework_1}.
% \vspace{-0.5cm}
\subsection{Sand Images Formulation}

The physical model that is widely used to describe the formation of an image suffered from light transmission hazed \cite{narasimhan2002vision, li2017haze} is often defined as follows:
{
\begin{equation}
\label{eq1}
    I(x) = J(x)t(x) + A(1 - t(x)),
\end{equation}
}
where $I(x)$ is the observed hazy image, $A$ is the global atmosphere light, and $t(x)$ is the medium transmission map, $J(x)$ is the haze-free image. Moreover, we have $t(x) = e^{-\beta d(x)}$ being the atmosphere scattering parameter and the scene depth, respectively. Based on the physical model, we construct a new dataset called SandPascal VOC ++, which contains three forms of sand.
\begin{figure}[!h]
    \centering
    \includegraphics[width=1\linewidth]{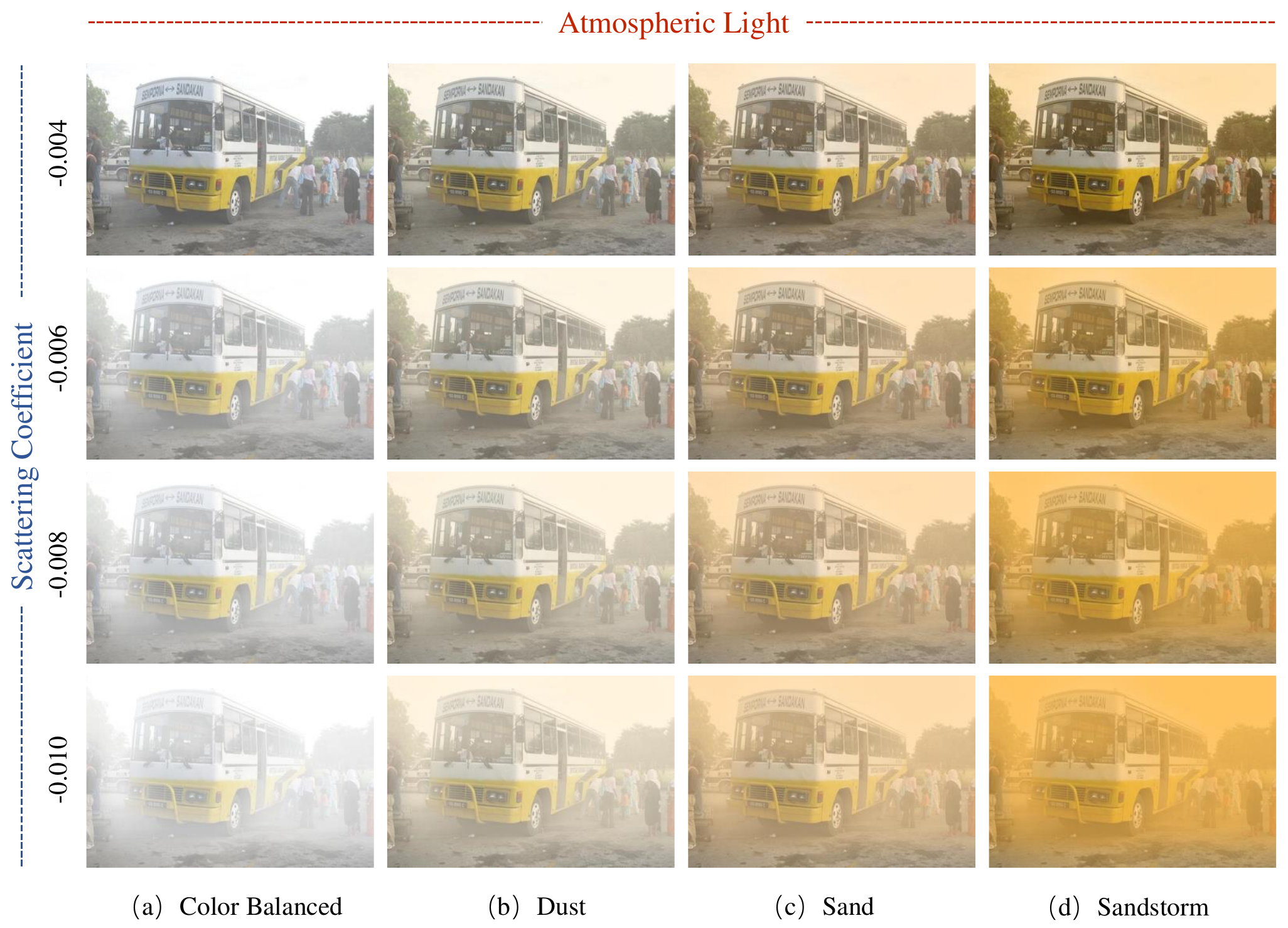}
    \caption{Analysis of the influence of different parameters in atmosphere scattering model.}
    \label{dataset}
\end{figure}

\textbf{Dataset generation}. In dusty weather, due to the different decay of $R$, $G$, and $B$ values, the degraded images have prior features such as offset, concentration and time. Considering the atmospheric light attenuation effect of the dust floating in the atmosphere on the $R$, $G$, and $B$ channels, the atmospheric light model belonging to the dust image was reconstructed according to the spatial distribution law. The mathematical expression of this model is:
{
\begin{equation}\label{eq3}
    \hat{A} = < A_{R}, k_{1}A_{R}+b_{1}, k_{2}A_{R}+b_{2} >,
\end{equation}
}
where $A_{G}=k_{1}A_{R}+b_{1}$, $A_{B}=k_{2}A_{R}+b_{2}$, $\hat{A}$ is the global color deviation value of the sand dust image, $k$ is the spatial distribution coefficient of the atmospheric light value of the three basic color spectrums, $b$ is the disturbance amount. Based on the formula \ref{eq3}, various dust images with different degrees of degradation can be synthesized from clear images through artificial algorithms. 
% \vspace{-0.5cm}
\subsection{CNN Block}

\begin{figure}[!h]
    \centering
    \includegraphics[width=0.8\linewidth]{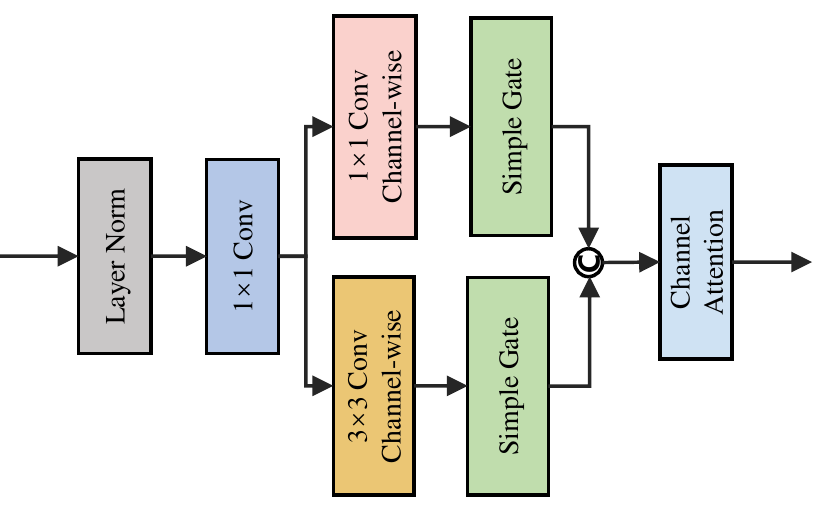}
    \caption{CNN Block}
    \label{CNN}
\end{figure}

Although CNN has achieved great success in the field of image restoration, the inherent properties of CNNs make the network's performance reach a bottleneck. Based on this, this paper introduces ViT to enable the network to achieve better generalization performance while maintaining the ideal characteristics of CNNs. Motivated by this, we propose a novel dual-branch gated attention residual module to obtain local feature information. This module is implemented by embedding simple gating and channel attention in a convolutional neural network. As shown in Figure \ref{CNN}, the degraded image is obtained through ResNet\cite{he2016deep} to obtain shallow network features of size $H_{0} \times W_{0} \times C_{0}$, which are respectively sent to the CNN branch and the Transformer branch. In order to keep the CNN features optimized in the CNN backbone, we add a downsampling module before each CNN Block to prevent overfitting. Both branches are deep-wise CNN to reduce the computational complexity; the convolution kernel sizes are 1 and 3, respectively, to achieve multi-scale feature extraction. To obtain a larger receptive field, we use simple channel attention at the end of each convolutional block, which fuses features at different scales to obtain richer feature representations.

% \cite{chen2022simple} reveals that the nonlinear activation functions, e.g., Sigmoid, ReLU, GELU, Softmax, Softmax, etc., are unnecessary: they could be replaced by multiplication or removed.  The essential operation in CNNs is the ‘convolution’ that provides local connectivity and translation equivariance.

%MobileNets\cite{howard2017mobilenets} efficiently encode local features by stacking depthwise and pointwise convolutions. The basic operation in CNNs is the ‘convolution’ that provides local connectivity and translation equivariance. While these properties bring efficiency and generalization to CNNs, they also cause two main issues. (a) The convolution operator has a limited receptive field, thus preventing it from modeling long-range pixel dependencies. (b) The convolution filters have static weights at inference, and thereby cannot flexibly adapt to the input content. To deal with the above-mentioned shortcomings, a more powerful and dynamic alternative is the self-attention (SA) mechanism [17,77,79,95] that calculates response at a given pixel by a weighted sum of all other positions.
% \vspace{-0.5cm}
\subsection{Transformer Block}

\begin{figure}[!h]
    \centering
    \includegraphics[width=1.0\linewidth]{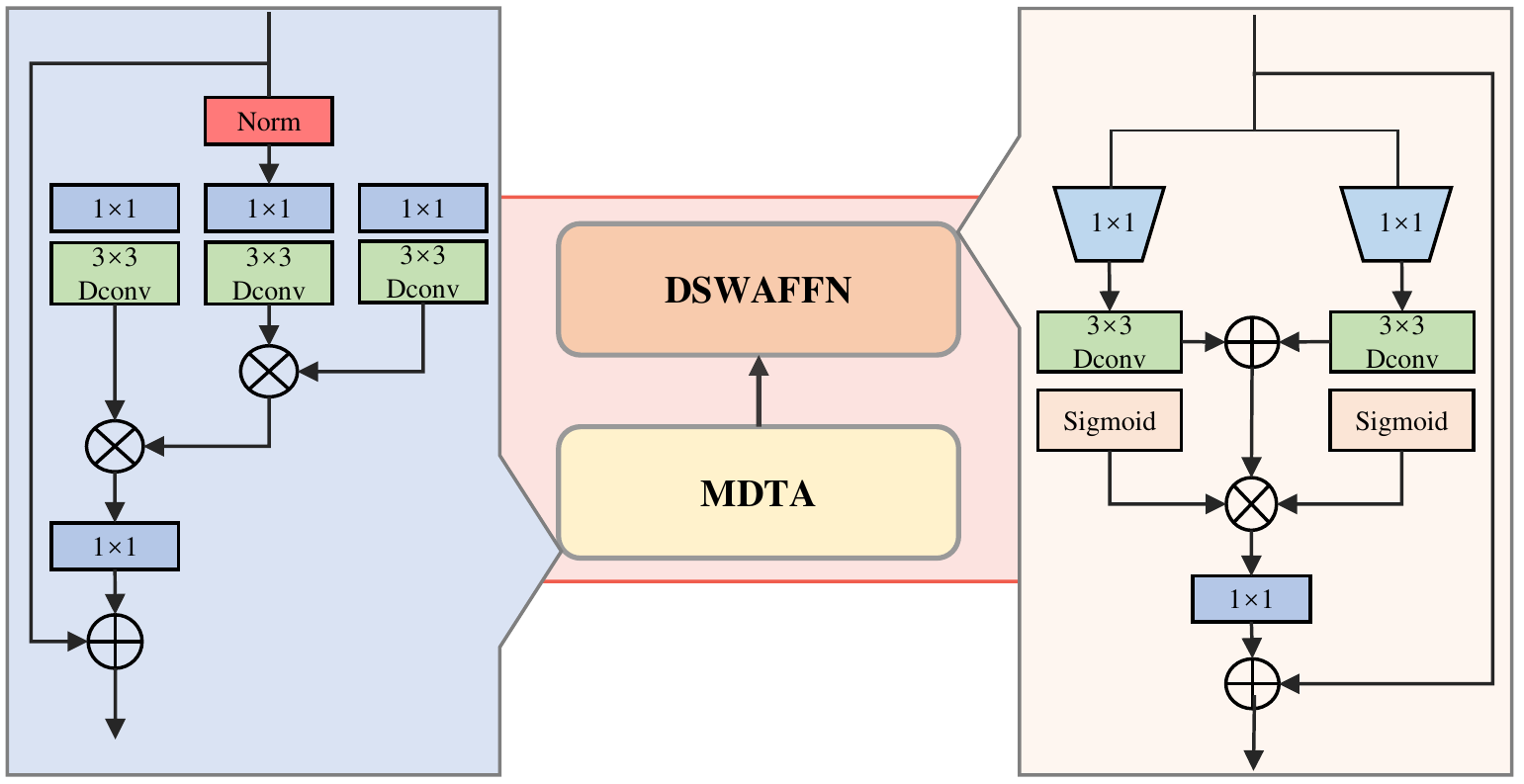}
    \caption{Transformer Block}
    \label{FFN}
\end{figure}

Due to the inductive bias of locality and weight sharing, the convolution operations demonstrate the intrinsic limitations in modeling the long-range dependency. The self-attention mechanism of Transfoemr itself makes it beneficial to capture long-term dependencies between input sequences.  In order to take advantage of the powerful representation ability of Transformer, we add the Transformer branch to the proposed method and propose a novel feed forward network (FFN) called Dual-path Shared Weight Attention FFN(DSWAFFN). Similar to the CNN branch, a dual-path form is also used to improve the fitting ability of this module. Meanwhile, to reduce the network complexity\cite{hu2018squeeze}, the two branches share the weight, and the sigmoid activation function is added to obtain the attention weight of each branch. For the self-attention part of the transformer branch, we use the same structure as in\cite{zamir2022restormer}, due to this can efficiently process high-resolution images while taking into account the capacity to handle global dependencies. A residual connection is at the end of this module, as shown in Figure \ref{FFN}, which allows the gradient to propagate effectively during backpropagation, thus avoiding the gradient-vanishing problem. Transformer stem aims to provide further guidance for global restoration with progressive features according to the convolution features. 

% \vspace{-0.5cm}

\subsection{Gate Fusion}

\begin{figure*}[!h]
    \centering
    \includegraphics[width=1.0\linewidth]{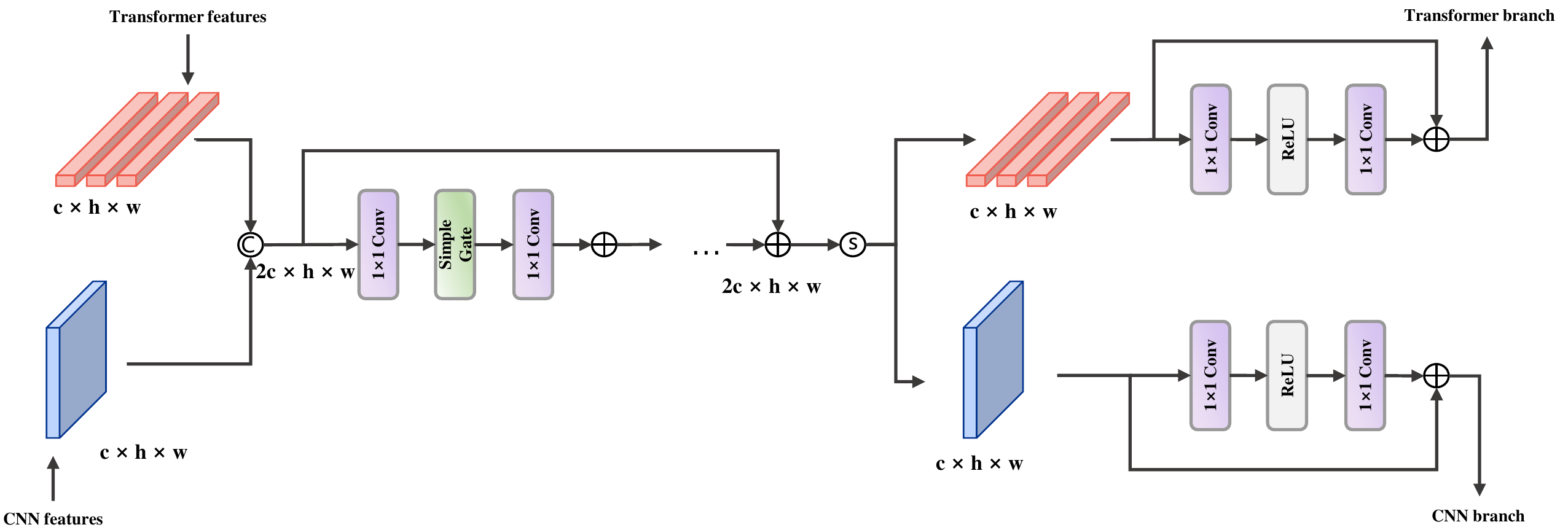}
    \caption{We concatenate features from two branches, bidirectionally transfer the mixed information to the original branches.}
    \label{fusion}
\end{figure*}

Using CNN or Transformer separately causes either local or global features to be neglected, which affects the model's performance. We propose a novel fusion module to fuse features from different branches by gating blocks to address the feature inconsistency between Transformer and CNN.  Specifically, feature maps are extracted from different levels of the CNN and Transformer branches and sent to the gate fusion module. We take a new step towards bridging the gap between CNNs and Transformer by presenting a new method to “softly” introduce a convolutional inductive bias into the ViT.
%To fully exploit the complementary features of Transform and CNN, we introduce gated fusion to aggregate multi-scale features carefully designed for image restoration to balance complexity and performance with spatial variance.

% \vspace{-1cm}
\begin{figure}[!h]
    \centering
    \includegraphics[width=1.0\linewidth]{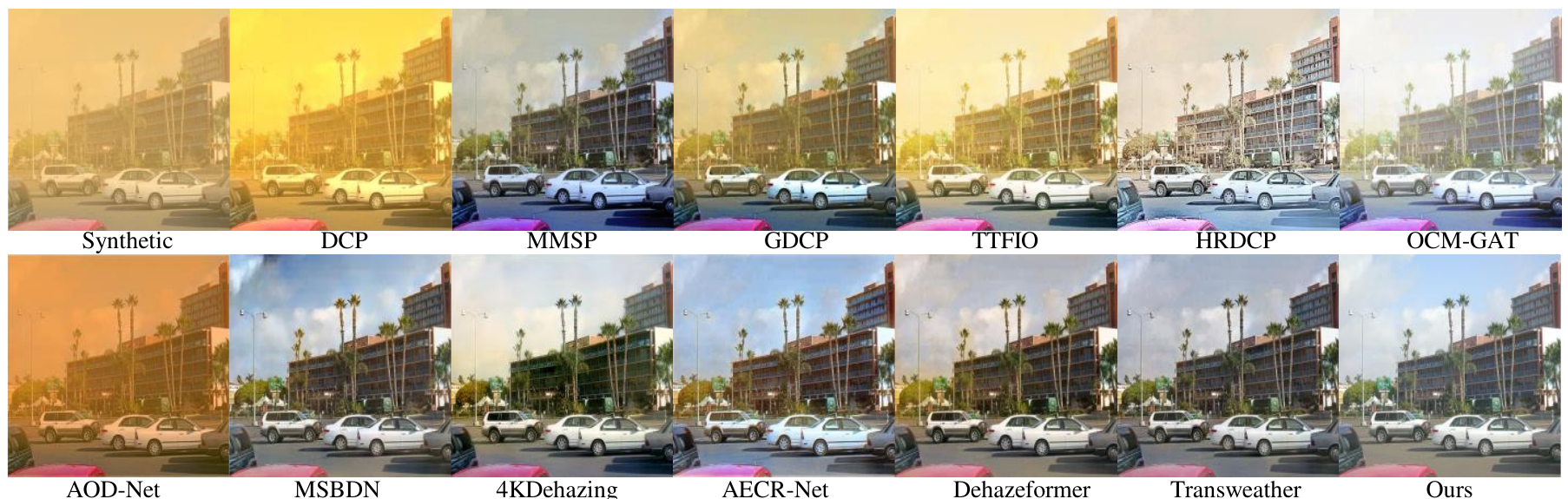}
    \caption{Image restoration results on a synthetic sand dust dataset.}
    \label{syntheticmethod}
\end{figure}

% \vspace{-1cm}

\begin{figure}[!h]
    \centering
    \includegraphics[width=1.0\linewidth]{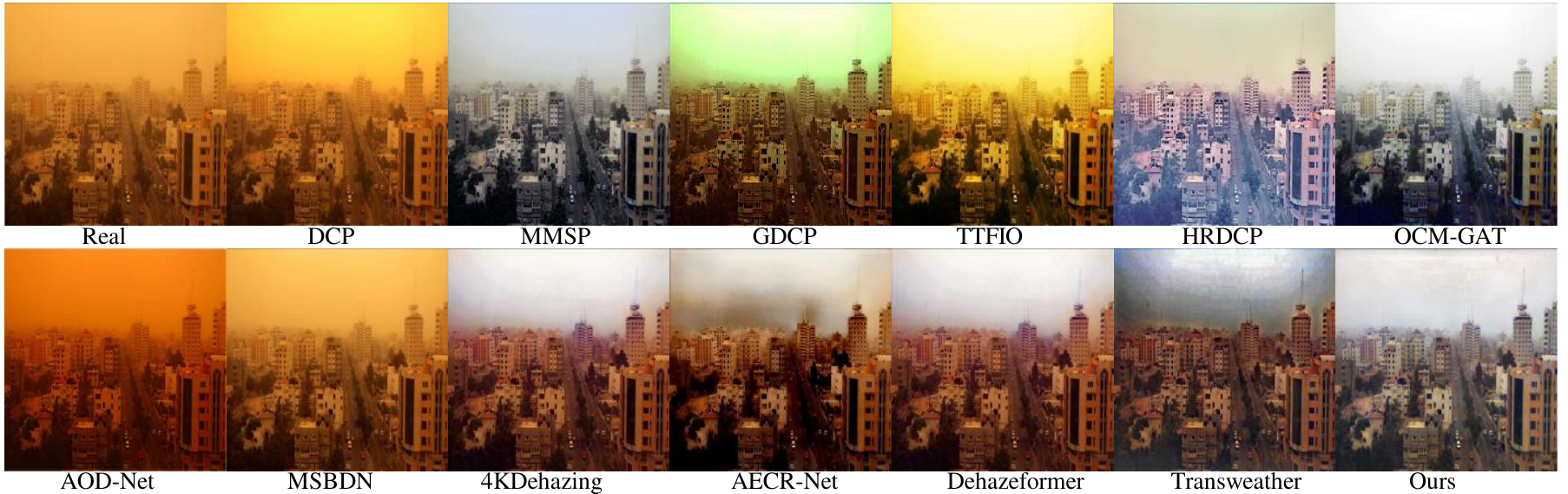}
    \caption{Image restoration results on the real world.}
    \label{realmethod}
\end{figure} 

As shown in Figure \ref{fusion}, we also integrate the idea of gating into the design of fusion module. A simple gate block is added to replace the ReLU activation function in the middle part of the residual block. At the end of the fusion module, we use the channel dimension convolution block for further feature extraction. Using the fusion module, we can ingeniously integrate the features extracted by CNN with the features extracted by Transformer rather than directly adding them together, which significantly reduces the performance of the whole model.
\vspace{-0.5cm}
\section{Experiments}

% \subsection{Experimental Dataset}
% We enrich the popular Pascal VOC named SandPascal VOC++ by considering atmospheric light value and concentration. Synthesize 18000 different types of sand dust images from Pascal VOC 2007 and 2012. Split it into a training set (containing 15000 images) and a testing set (containing 3000 images). We collect real sand dust images by shooting the network to evaluate the generalization performance of the network in the real domain.
% \vspace{-0.5cm}
\subsection{Experimental results}
This paper compares the sand dust image restoration method based on traditional prior and the haze image restoration method based on deep learning. For fairness, we retrain all networks on SandPascal VOC++. For the comparative experiment section, we have provided some visual effect comparison pictures in \ref{syntheticmethod} and \ref{realmethod}. We obtain the best results of the proposed method by training SandPascal VOC ++, combined with our proposed CNN branch, Transformer branch, and fusion module. We compare it with twelve State-of-the-Art methods and re-evaluate all methods, as shown in Table \ref{comparionTable}. Meanwhile, the visualized images in \ref{syntheticmethod} and \ref{realmethod} match well with the quantitative results, showing our proposed method's favorable image restoration capability. 
% For other methods of comparison, to be fair, we retrain them using our newly proposed dataset SandPascal VOC++. For our proposed method, during training, the synthesized sand images and the corresponding label images are resized to $256 \times 256$ and normalized to $(0, 1)$. The batch size use for training is set to 8, and we train our model with AdamW optimizer for total 500 epochs with the initial learning rate $1e^{-4}$.
% \vspace{-0.5cm}
\subsection{Ablation Study}
For the ablation experiments section, as shown in Table \ref{ablaExp}, we illustrate the importance of individual components of our model. The model's performance is greatly reduced when there are only Transformer and CNN branches. Meanwhile, when the output of these two is simply added, the model's performance will also be affected. Finally, the three parts we proposed are integrated. That is, our proposed method achieves the best results, which shows that all parts of the network are of importance.

% \vspace{-1em}

\begin{table}[!h]

\setlength{\abovecaptionskip}{0cm}  %段前
\setlength{\belowcaptionskip}{-0.2cm} %段后

\centering
\caption{Comparative results on synthetic images and real images, all models are trained on our proposed dataset SandPascal VOC++.}
\label{comparionTable}
\resizebox{\linewidth}{!}{
\begin{tabular}{c|cc|ccc}
\hline
\multirow{2}{*}{Method} & \multicolumn{2}{c|}{SandPascal VOC++} & \multicolumn{2}{c}{Real} \\ \cline{2-6} 
                        & PSNR↑              & SSIM↑            & NIQE↓     &NIMA↑ & User Study↑    \\ \hline
DCP\cite{5567108}                     & 17.925             & 0.855            & 3.361     & 3.998 & 2.6               \\
MMSP\cite{fu2014fusion}                    & 17.772             & 0.843            & 2.926     & 3.943 & 7.3              \\
GDCP\cite{peng2018generalization}                    & 13.669             & 0.752            & 3.255     & 3.890 &3.9               \\
TTFIO\cite{al2016visibility}                   & 15.324             & 0.789            & 3.476     & 3.924 & 4.3              \\
HRDCP\cite{shi2019let}                   & 12.212             & 0.683            & 4.131     & 3.884 &5.8               \\
OCM-GAT\cite{yang2020visibility}                 & 16.001             & 0.800            & 3.020  &3.850     & 7.5              \\
AOD-Net\cite{li2017all}                 & 17.799             & 0.846            & 3.175      & 3.998 & 1.6              \\
MSBDN\cite{dong2020multi}                   & 29.033             & 0.918            & 2.988     & 3.726 & 3.6              \\
4KDehazing\cite{2021Ultra}              & 28.215             & 0.912            & 3.447     &3.852 &  5.3             \\
AECR-Net\cite{9578448}                & 27.222             & 0.907            & 3.146     & 3.886 & 4.2               \\
Dehazeformer\cite{song2022vision}            & 30.123             & 0.929            & 2.913      & 3.923 & 5.8              \\
Transweather\cite{2021TransWeather}            & 30.621             & 0.928            & 3.019     & 4.040 &  4.4              \\
Sand\_images            & -                  & -                & -     & -  & -               \\ \hline
\hline \gr Ours                    & 34.150             & 0.952            & 2.795     &4.078 & 8.4               \\ \hline
\end{tabular}
}
\end{table}

% \vspace{-2em}

\begin{table}[!h]
\caption{Ablation experiments of proposed method.}
\label{ablaExp}
% \resizebox{\linewidth}{!}{
\centering
\begin{tabular}{c|cc}
\hline
                   & PSNR↑   & SSIM↑  \\ \hline
only Transformer Branch & 31.426 & 0.941 \\
only CNN Branch         & 20.449 & 0.826 \\
Transformer + CNN  & 30.986 & 0.941 \\
Trans. + CNN + SK Fusion        & 31.667 & 0.948 \\ \hline              
\hline \gr Trans. + CNN + Gate Fusion        & 34.150 & 0.952  \\ \hline
\end{tabular}
% }
\end{table}

% \vspace{-3em}
% \vspace{-0.5cm}
\section{Conclusion}

In this work, we explore the visual effects of SandFormer, formulating a synthetic dataset with simultaneously fugitive dust, sand, and sandstorms. And proposed a sand dust image restoration network based on CNN and Transformer (called SandFormer), which combines the advantages of CNN in local feature recovery and Transformer in global perception, and obtains the final clear image through gate fusion. Extensive experiments show that our method outperforms state-of-the-art sand dust image restoration methods both quantitatively and qualitatively in dust scenes. In the future, we will incorporate high-level vision tasks.
 
\vfill\pagebreak

% \section{REFERENCES}
% \label{sec:refs}

% List and number all bibliographical references at the end of the
% paper. The references can be numbered in alphabetic order or in
% order of appearance in the document. When referring to them in
% the text, type the corresponding reference number in square
% brackets as shown at the end of this sentence \cite{C2}. An
% additional final page (the fifth page, in most cases) is
% allowed, but must contain only references to the prior
% literature.

% \bibliographystyle{IEEEbib}
% \bibliography{main}

\end{document}